\documentclass[letterpaper, 10 pt, conference]{ieeeconf}  

\IEEEoverridecommandlockouts                              

\pdfoutput=1

\usepackage{hyperref}  
\usepackage{amsmath,amssymb,amsfonts}
\usepackage{algorithmic}
\usepackage{graphicx}
\usepackage{textcomp}
\usepackage{xcolor}
\usepackage{biblatex}
\usepackage{array}
\usepackage{booktabs}
\usepackage{multirow}
\usepackage[utf8]{inputenc}
\usepackage{amsmath}
\usepackage{dsfont}
\usepackage[american]{babel}
\usepackage{microtype}

\title{\LARGE \bf
Learning an Adaptive Fall Recovery Controller for Quadrupeds on Complex Terrains
}

\author{Yidan Lu$^{*}$, Yinzhao Dong$^{*}$, Ji Ma, Jiahui Zhang, Peng Lu$^{\dagger}$
\thanks{$^{*}$ Equal Contribution.}
\thanks{$^{\dagger}$Corresponding author: \url{lupeng@hku.hk} }
\thanks{The authors are with the Adaptive Robotic Controls Lab (ArcLab), Department of Mechanical Engineering, The University of Hong Kong, Hong Kong SAR, China, \url{ydlu@connect.hku.hk}, \url{dongyz@connect.hku.hk},  \url{maji@connect.hku.hk}, \url{holmesz@connect.hku.hk}}
\thanks{This work was supported by the General Research Fund under Grant 17204222, and in part by the Seed Fund for Collaborative Research and General Funding Scheme-HKU-TCL Joint Research Center for Artificial Intelligence.}
}

\begin{document}

\maketitle
\thispagestyle{empty}
\pagestyle{empty}

\begin{abstract}
Legged robots have shown promise in locomotion complex environments, but recovery from falls on challenging terrains remains a significant hurdle. This paper presents an Adaptive Fall Recovery (AFR) controller for quadrupedal robots on challenging terrains such as rocky, breams, steep slopes, and irregular stones. We leverage deep reinforcement learning to train the AFR, which can adapt to a wide range of terrain geometries and physical properties. Our method demonstrates improvements over existing approaches, showing promising results in recovery scenarios on challenging terrains. We trained our method in Isaac Gym using the Go1 and directly transferred it to several mainstream quadrupedal platforms, such as Spot and ANYmal. Additionally, we validated the controller's effectiveness in Gazebo. Our results indicate that the AFR controller generalizes well to complex terrains and outperforms baseline methods in terms of success rate and recovery speed.
\end{abstract}

\section{INTRODUCTION}
Legged robots have made significant strides in locomotion capabilities, demonstrating impressive performance in tasks such as dynamic walking, running, and even complex maneuvers like backflips \cite{lee2020learning}, \cite{garcia2021time}. However, the ability to recover from falls, especially on challenging and unpredictable terrains, remains a critical challenge in the field of legged robotics. While substantial progress has been made in recovery strategies for flat or moderately uneven surfaces \cite{lee2019robust}, \cite{nahrendra2023robust}, the problem of robust recovery on highly irregular terrains – such as rocky landscapes, steep inclines, or complex gaps – has received limited attention.

The ability to recover from falls on challenging terrains is crucial for deploying legged robots in real-world scenarios, including search and rescue operations, planetary exploration, and industrial inspection tasks. In these challenging applications, robots must locomotion unpredictable and potentially hazardous environments where falls are likely to occur. A robust recovery ability would significantly enhance the autonomy and reliability of legged robots in such situations. However, achieving effective recovery for legged robots in complex terrains is a significant challenge. In addition to the inherent instability of legged designs, recovery is complicated by unpredictable environmental factors such as uneven surfaces and dynamic obstacles. External perturbations, such as pushes or slips, further complicate stability.



\subsection{Model-Based Fall Recovery for Quadruped Robots}
In the past decade, significant advancements have been made in continuous locomotion for robots \cite{luo2024moral},  \cite{zhuang2023robot}. How to maintain balance and achieve rapid recovery from slipping has been one of the key metrics for the continuous locomotion of quadrupedal robot controllers  \cite{khorram2015balance}, \cite{focchi2018slip}. 
Currently, most model-based work focuses on fall prediction and falling control. For instance, Yang et al. \cite{yang2018falling} proposed a fall prediction model that can identify and anticipate potential fall situations for the robot during locomotion. Rajesh et al. \cite{mordatch2012discovery} introduced an innovative framework that enables the robots to actively manage energy during falls and generate rolling trajectories in real-time, ensuring that the robots can dynamically adapt to various fall scenarios. 

However, in extreme or complex natural environments, robots still face the inevitability of falling. A major challenge in current research lies in developing adaptive controllers for robots to effectively recover from falls, allowing them to resume movement or efficiently complete tasks. However, model-based methods are often inadequate for these dynamic tasks. For example, Mordatch et al. \cite{mordatch2012discovery} proposed a framework that optimizes automatic recovery through contact invariance, but the reliance on predefined potential contact points limits the exploration of flexible behaviors. In addition, classical optimal control algorithms may struggle to model non-smooth dynamic contact scenarios, as they often rely on simplified models and predefined contact sequences.  

\begin{figure*}[h]
   \vspace{-1.5em}
   \centering\includegraphics[width=0.90\textwidth,height=0.30\textwidth]{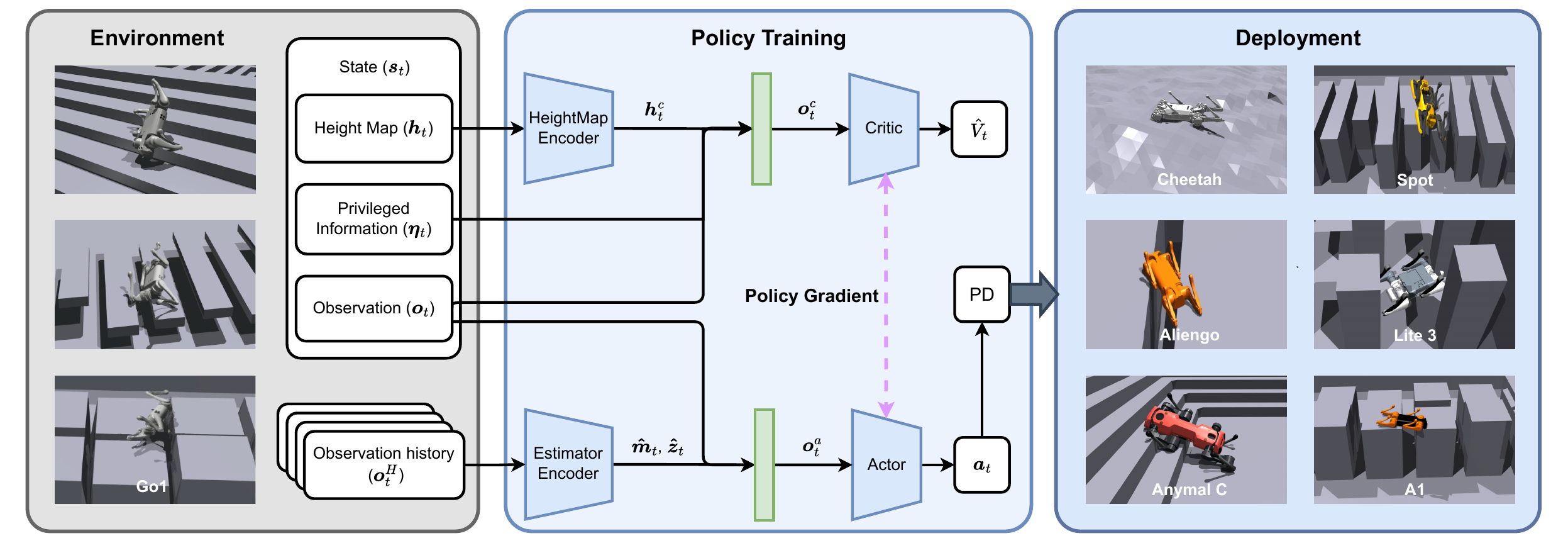}
   \caption{Overview of the AFR framework. Trained on Go1 could be deployed on multiple platforms.}
   \label{pics_framework}
   \vspace{-1.5 em}
\end{figure*}
\subsection{Learning-Based Fall Recovery for  Robots}
Model-free Deep Reinforcement Learning (DRL) has been widely utilized for fall recovery tasks in quadrupedal robots, achieving significant performance. This is because DRL is suited for handling dynamic continuous decision-making challenges, and it also eliminates the need for precise modeling of complex collision scenarios through extensive interactions with the environment. Early work on robot recovery focused primarily on flat or nearly flat surfaces  \cite{lee2019robust}, \cite{hwangbo2019learning}. Many approaches have focused on moderately uneven surfaces. For example, The DribbleBot system \cite{ji2023dribblebot} demonstrated the ability to recover and continue task execution on various outdoor surfaces, including sand, gravel, and snow. However, their approach was not explicitly designed for highly irregular or geometric terrains like stairs. \cite{smith2022legged}, \cite{li2024dynamic} also use DRL for recovery on uneven terrain, but these method was limited to moderately uneven surfaces and did not address the complexities of highly challenging terrains.  Recently, \cite{nahrendra2023robust} presented a method to learn the terrain imagination for robust fall recovery, which can adapt to various terrains, including sponges, irregular bumps, piles of boxes, and stairs.

Despite these advancements, there remains a significant gap in the literature concerning recovery strategies for highly challenging and irregular terrains, such as complex rocky, cantilever beams, and even on irregular structures like piles or clusters of obstacles. Therefore, this research aims to fill this gap by developing a learning-based approach capable of performing fall recovery across various complex geometries and adapting to the diverse physical properties encountered in different scenarios, such as friction.




\subsection{Contribution}
In this paper, we propose an Adaptive Fall Recovery (AFR) controller for quadrupedal recovery skills to enhance the adaptability and resilience of quadrupedal robots, enabling them to perform effective fall recovery in complex scenarios that are often unpredictable and difficult to balance. The key contributions of this work can be listed as follows:

\begin{itemize}
 \item A novel DRL framework for training recovery controllers that can generalize to a wide range of terrain geometries and physical properties.

 \item A versatile fall recovery policy that can be deployed on various quadrupedal robot platforms, demonstrating its broad applicability and adaptability to different robot configurations.

 \item Extensive experimental validation via sim-to-sim transfer, demonstrating the effectiveness and generalization capabilities of our approach across different simulation platforms and on previously unseen terrain types. 
\end{itemize}
\section{METHOD}

\subsection{Problem Formulation}
We formulate the problem of quadrupedal fall recovery as a Partially Observable Markov Decision Process (POMDP) \cite{spaan2012partially}, denoted as a 7-tuple $\mathcal{M} = \{ \mathcal{S}, \mathcal{O}, \mathcal{A}, \mathcal{R}, \mathcal{T}, \Omega, \gamma\}$, where $\mathcal{S}$, $\mathcal{O}$ and $\mathcal{A}$ are the set of states, observations, and actions, respectively. For each state $\boldsymbol{s_t} \in \mathcal{S}$, the learning agent interacts with the environment with an action $\boldsymbol{a_t} \in \mathcal{A}$ and receives a reward $\mathcal{R}(\boldsymbol{s_t}, \boldsymbol{a_t})$, leading to the transition of the environment to the next state $\boldsymbol{s}_{t+1}$ with the probability $\mathcal{T}(\boldsymbol{s}_{t+1}|\boldsymbol{s_t}, \boldsymbol{a_t})$. Meanwhile, the observation $\boldsymbol{o}_{t+1} \in \mathcal{O}$ depends on the new state $\boldsymbol{s}_{t+1}$ and the action $\boldsymbol{a_t}$ with a conditional probability $\Omega (\boldsymbol{o}_{t+1}|\boldsymbol{s}_{t+1}, \boldsymbol{a_t})$. The objective is to determine the optimal policy $\pi^{*}$ that maximizes the accumulated rewards of this POMDP $\mathcal{M}$, considering a discount ratio $\gamma$, i.e.: $J_{\mathcal{M}}(\boldsymbol{\pi}) = \mathbb{E}_{\pi}\left[\sum_{t=0}^{T-1}(\gamma^{t}\mathcal{R}(\boldsymbol{s}_t, \boldsymbol{a}_t))\right].$



\noindent\textbf{Observation and State Space} The robot observation $\boldsymbol{o}_{t}$ in our control scheme can be collected from the proprioceptive sensors, including body angular velocity $\boldsymbol{\omega}_t\in\mathbb{R}^{3}$, projected gravity $\boldsymbol{g}_{t}\in\mathbb{R}^{3}$, joint angles $\boldsymbol{q}_{t}\in\mathbb{R}^{12}$, joint angular velocities $\boldsymbol{\dot{q}}_{t}\in\mathbb{R}^{12}$, and the action of the last step $\boldsymbol{a}_{t-1}\in\mathbb{R}^{12}$, which can be defined as: $\boldsymbol{o}_{t} = \left( \boldsymbol{\omega}_t,\ \boldsymbol{g}_{t},\ \boldsymbol{q}_{t},\ \boldsymbol{\dot{q}}_{t}, \ \boldsymbol{a}_{t-1} \right)$

The state $\boldsymbol{s}_t = (\boldsymbol{o}_t, \boldsymbol{h}_t, \boldsymbol{\eta}_t)$ includes the standard observation $\boldsymbol{o}_t$, terrain height map $\boldsymbol{h}_t \in \mathbb{R}^{187}$, and privileged information $\boldsymbol{\eta}_t$. $\boldsymbol{\eta}_t = ( \boldsymbol{m} \in \mathbb{R}^4, \boldsymbol{K}_\text{PD} \in \mathbb{R}^{12}, \boldsymbol{p}_\text{CoM} \in \mathbb{R}^3, \mu \in \mathbb{R}, \boldsymbol{c} \in \mathbb{R}^4, \boldsymbol{f} \in \mathbb{R}^{12})$ represents masses (body, hip, thigh, calf), PD gains, CoM position, friction coefficient, foot contact states, and foot forces, respectively.

\noindent \textbf{Action Space} 
The action $\boldsymbol{a}_t\in\mathbb{R}^{12}$ represents the desired increment of the joint angle w.r.t the initial pose $\boldsymbol{\mathring{q}}$, i.e. $\boldsymbol{q}^{*}_{t} = \boldsymbol{\mathring{q}} + \boldsymbol{a}_{t}$. The final desired angle $\boldsymbol{q}^{*}_{t}$ is tracked by the torque generated by the joint-level proportional–derivative (PD) controller of the joint-level actuation module in the simulator, i.e. $\boldsymbol{\tau} = \boldsymbol{k}_p \cdot (\boldsymbol{q}^{*}_{t} - \boldsymbol{q}_t) + \boldsymbol{k}_d \cdot ( - \boldsymbol{\dot{q}}_{t})$.

\subsection{Adaptive Fall Recovery Controller}
As show in Fig. \ref{pics_framework}, we employ the Proximal Policy Optimization (PPO) algorithm \cite{schulman2017proximal} to train an Adaptive Fall Recovery (AFR) controller, enabling the quadrupedal robots to perform effective fall recovery in complex scenarios. The AFR controller consists of four sub-networks: Estimator Encoder, HeightMap Encoder, Actor Network, and Critic Network. These components work together to achieve real-time adaptation to various terrains and robot dynamics. Next, each part of the AFR will be discussed in detail.

\begin{table*}[htbp]\footnotesize
\caption{Reward terms for quadrupedal recovery}
\vspace{-1.0em}
\centering
\small
\begin{tabular}{cccc}
\toprule
Category & Reward & Equation & Weight ($w_i$) \\
\midrule
\multirow{3}{*}{Orientation \& Posture}
 & Base Orientation & $ \boldsymbol{\boldsymbol{g}}_{xy}^2$ & -0.5 \\[1ex]
 & Upright Orientation & $\exp(-\frac{(g_z + 1)^2}{2\epsilon^2})$ & 6.0 \\
 & Target Height Alignment  & $\exp(-(h_{target} - h_{current})^2)$ & 1.0 \\
\midrule
\multirow{2}{*}{Stability \& Configuration}
 & Feet on Ground & $\sum_{i=1}^4 \mathds{1}_{f_{contact_i}}$ & 0.3 \\
 & Target Posture & $\exp(-(\boldsymbol{q} - \boldsymbol{q}_{stand})^2)$ if $|g_z + 1| < \epsilon$ & 4.0 \\
\midrule
\multirow{4}{*}{Motor Control}
 & Action & $ \boldsymbol{a}^2_t$ & $-1.0\text{e-}2$ \\
 & Joint Torques & $ \boldsymbol{\tau}^2$ & $-5.0\text{e-}4$ \\[1ex]
 & Joint Acceleration & $\ddot{\boldsymbol{q}}^2$ & $-2.5\text{e-}6$ \\
 & Joint Velocity & $ \dot{\boldsymbol{q}}^2$ & $-1.0\text{e-}2$ \\
\midrule
\multirow{2}{*}{Safety}
 &Base-Ground Contact Penalty & $ \mathds{1}_{b_{contact}}$ & -0.2 \\
 &Position Limits & $ \sum_{i=1}^{12} \mathds{1}_{q_i>q_{\max}||q_i<q_{\min}}$ & -1.0 \\
\midrule
\multirow{3}{*}{Motion Smoothness}
 & Angular Velocity Limit& $\max(|\dot{\boldsymbol{q}}| - 0.8, 0)$ & -0.1 \\
 & Action Smoothing  & $(\boldsymbol{a}_t - \boldsymbol{a}_{t-1})^2 + (\boldsymbol{a}_t - 2\boldsymbol{a}_{t-1} + \boldsymbol{a}_{t-2})^2$ & -0.05 \\
\bottomrule
\end{tabular}
\vspace{-1.5em}
\label{tab:reward_terms}
\end{table*}
\subsubsection{Estimator Encoder}

The Estimator Encoder $E_\phi(\boldsymbol{o}_t^H)$  parameterized by $\phi$ processes $H$ consecutive observation frames to estimate robot mass distribution and extract temporal features:
\begin{equation}
\boldsymbol{\hat{m}}_t, \boldsymbol{\hat{z}}_t = E_\phi(\boldsymbol{o}_t^H)
\end{equation}
where $\boldsymbol{o}_t^H= \left[ \boldsymbol{o}_{t-H+1}, ..., \boldsymbol{o}_{t-1}, \boldsymbol{o}_{t} \right]$ is a temporal observation sequence ($H=5$ in this task). $\boldsymbol{\hat{m}}_t \in \mathbb{R}^4$ represents the estimated masses of critic links (including base, hip, thigh, and calf), and $\boldsymbol{\hat{z}}_t$ is a latent vector encoding temporal information. Meanwhile, the training of the estimator net is not independent of the training of the actor, the parameters are also updated by a regression loss $loss_{reg}= MSE(\hat{\boldsymbol{m}}_t, \boldsymbol{m})$ to reduce the gap with ground truth of the link masses.
\subsubsection{HeightMap Encoder}

The HeightMap Encoder $H_\xi$  parameterized by $\xi$ compresses the terrain height map information:
\begin{equation}
\boldsymbol{h}_t^c = H_\xi(\boldsymbol{h}_t)
\end{equation}
where $\boldsymbol{h}_t \in \mathbb{R}^{187}$ is the original terrain height map, and $\boldsymbol{h}_t^c$ is the compressed height feature. This step allows the critic to efficiently process the high-dimensional terrain information.

\subsubsection{Actor Network}

The actor network $\pi_\theta(\boldsymbol{a}_t|\boldsymbol{o}^a_t)$ parameterized by $\theta$ represents the policy:
\begin{equation}
\pi_\theta(\boldsymbol{a}_t|\boldsymbol{o}^a_t) = \mathcal{N}(\mu_\theta(\boldsymbol{o}^a_t), \sigma_\theta(\boldsymbol{o}^a_t))
\end{equation}
where $\mu_\theta$ and $\sigma_\theta$ are the mean and standard deviation of the action distribution, respectively. The input of actor $\boldsymbol{o}^a_t = (\boldsymbol{o}_t, \boldsymbol{\hat{m}}_t, \boldsymbol{\hat{z}}_t)$ consists of the current observation $\boldsymbol{o}_t$ and the outputs of the Estimator Encoder ($\boldsymbol{\hat{m}}_t$ and $\boldsymbol{\hat{z}}_t$).

\subsubsection{Critic Network}

The critic network $V_\psi(\boldsymbol{o}^c_t)$, parameterized by $\psi$, estimates the value function:
\begin{equation}
V_\psi(\boldsymbol{o}^c_t) \approx \mathbb{E}_\pi\left[\sum_{k=0}^{\infty} \gamma^k r(\boldsymbol{o}^c_{t+k}, \boldsymbol{a}_{t+k}) | \boldsymbol{o}^c_t\right]
\end{equation}
where $\boldsymbol{o}^c_t = (\boldsymbol{o}_t, \boldsymbol{\eta}_t, \boldsymbol{h}_t^c)$ is the input of critic, consisting of the standard observation $\boldsymbol{o}_t$, privileged information, and compressed height feature $\boldsymbol{h}_t^c$ from the HeightMap Encoder. $\gamma$, $r$, and $\mathbb{E}_\pi$ denote the discount factor, the reward function, and the expected value under the current policy $\pi$.

\subsection{Reward Function Design}
The reward functions are inherited from our previous works ~\cite{luo2024moral}, as shown in Table \ref{tab:reward_terms}. It can balance multiple objectives to encourage the robot to achieve and maintain a stable standing posture while minimizing energy consumption and avoiding harmful configurations, as follows: 

\begin{itemize}
    \item \textbf{Orientation and Posture}: Encourages an upright posture and proper height, minimizing tilt and ensuring vertical alignment.
    
    \item \textbf{Stability and Configuration}: Promotes stable foot placement and guides the robot to a desired standing configuration.
    
    \item \textbf{Motor Control}: Manages movements and energy use by penalizing excessive actions and forces.
    
    \item \textbf{Safety}: Discourages damaging configurations with penalties for ground contact and joint limits.
    
    \item \textbf{Motion Smoothness}: Promotes fluid movements by penalizing rapid changes and excessive velocities.
\end{itemize}
    
    
    
    

This comprehensive reward function design enables the robot to learn a recovery strategy that is stable, efficient, and safe, while also producing smooth and natural movements.



\subsection{Curriculum Learning}
We adopted a curriculum learning \cite{wang2021survey} to facilitate progressive learning of recovery abilities across challenging terrains. Each terrain type incorporates randomization in obstacle placement, sizes, and gaps to improve robustness and prevent overfitting. The difficulty of each terrain can be continuously adjusted by modifying parameters such as obstacle density, height variations, and gap sizes, as follows: 

\begin{itemize}
    \item \textbf{Slope}: Inclined rough surfaces with angles ranging from 0° to 45°, challenging the robot's balance and traction.
    \item \textbf{Discrete Obstacles}: Randomly placed obstacles with heights varying from 0.05 m to 0.3 m.
    \item \textbf{Stairs}: Stair heights range from 0.05 m to 0.25 m, with depths from 0.2 m to 0.5 m. 
    \item \textbf{Single Gaps}: Platforms feature gaps ranging from 0.1 m to 0.5 m wide.
    \item \textbf{Air Beams}: Narrow, elevated beams (0.1 m to 0.3 m wide, 0.1 m to 0.5 m high) requiring precise limb placement and balance for recovery.
    \item \textbf{Beams}: Multiple parallel beams of varying widths and spacings, testing the robot's ability to recover on discontinuous surfaces.
    \item \textbf{Sparse Stones}: Irregularly placed stepping stones with sizes varying from 0.15 m to 0.25 m and gaps between stones ranging from 0.3 m to 0.5 m, challenging the robot to find stable footholds during recovery.
    \item \textbf{Dense Stones}: Closely packed stones with sizes varying from 0.15 m to 0.40 m and gaps ranging from 0.10 m to 0.30 m, simulating highly uneven terrain for complex recovery scenarios.
\end{itemize}


\section{Experiments}
\subsection{Implementation Details}
Our controller was trained in NVIDIA Isaac Gym \cite{makoviychuk2021isaac}, leveraging its capability for efficient parallel simulations. We trained the policy by PPO \cite{schulman2017proximal} with 4,096 parallel domain-randomized Unitree Go1 quadrupedal robot. Episodes begin with a random supine position and terminate after 350 timesteps or upon maintaining a stable standing posture for 100 consecutive timesteps. The networks were optimized by the Adam \cite{kingma2014adam}. Training was conducted on a workstation with an Intel Core i7-13700K CPU, 32 GB RAM, and an NVIDIA RTX 4070 GPU. To enhance policy robustness, we implemented domain randomization as shown in Table \ref{tab:domain_randomization}. 
\vspace{-1.0em}
\begin{table}[h]\footnotesize
\centering
\caption{Domain randomization ranges applied in simulation.}
\vspace{-1.0em}
\begin{tabular}{ccc}
\hline
Parameter & Range & Units \\
\hline
Payload & $[-2.5, 2.5]$ & kg \\
$K_p$ factor & $[0.9, 1.1]$ & Nm/rad \\
$K_d$ factor & $[0.9, 1.1]$ & Nms/rad \\
Motor strength factor & $[0.9, 1.1]$ & Nm \\
COM shift & $[-50, 50]$ & mm \\
Trunk Mass   & $[4.0, 28.0]$ & kg \\
Hip Mass   & $[0.3, 0.7]$ & kg \\
Thigh Mass   & $[0.4, 4.0]$ & kg \\
Calf Mass   & $[0.1, 0.8]$ & kg \\
\hline
\end{tabular}
\label{tab:domain_randomization}
\end{table}
\vspace{-2.0em}
\subsection{Compared Methods}

\begin{figure}[h]
   \centering
   \includegraphics[width=0.45\textwidth, height=0.2\textwidth]{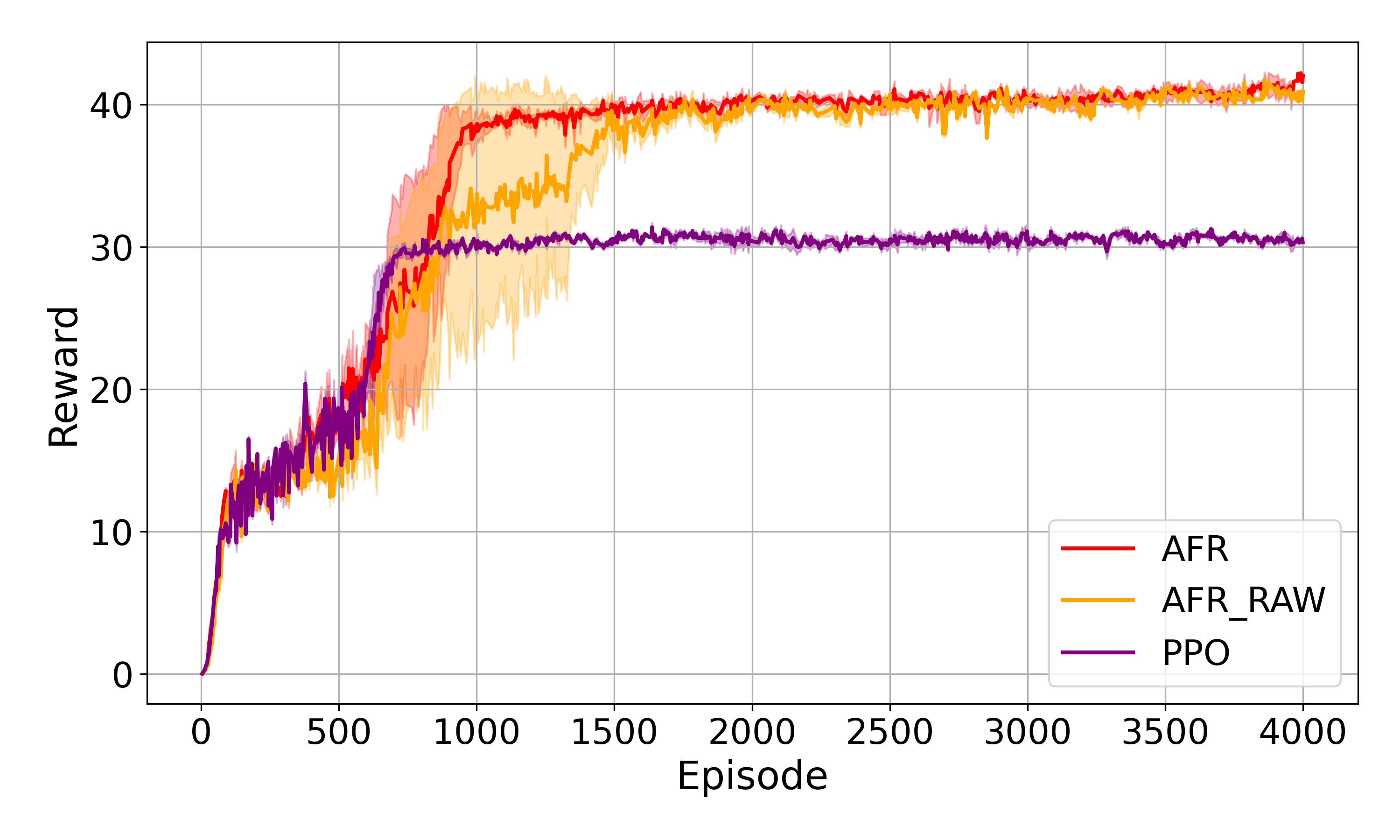}
   \caption{Total reward comparison among AFR, AFR\textunderscore RAW and PPO.}
   \label{fig:overall_reward}
   \vspace{-0.5em}
\end{figure}

\begin{figure}[h]
   \centering
   \includegraphics[width=0.45\textwidth, height=0.2\textwidth]{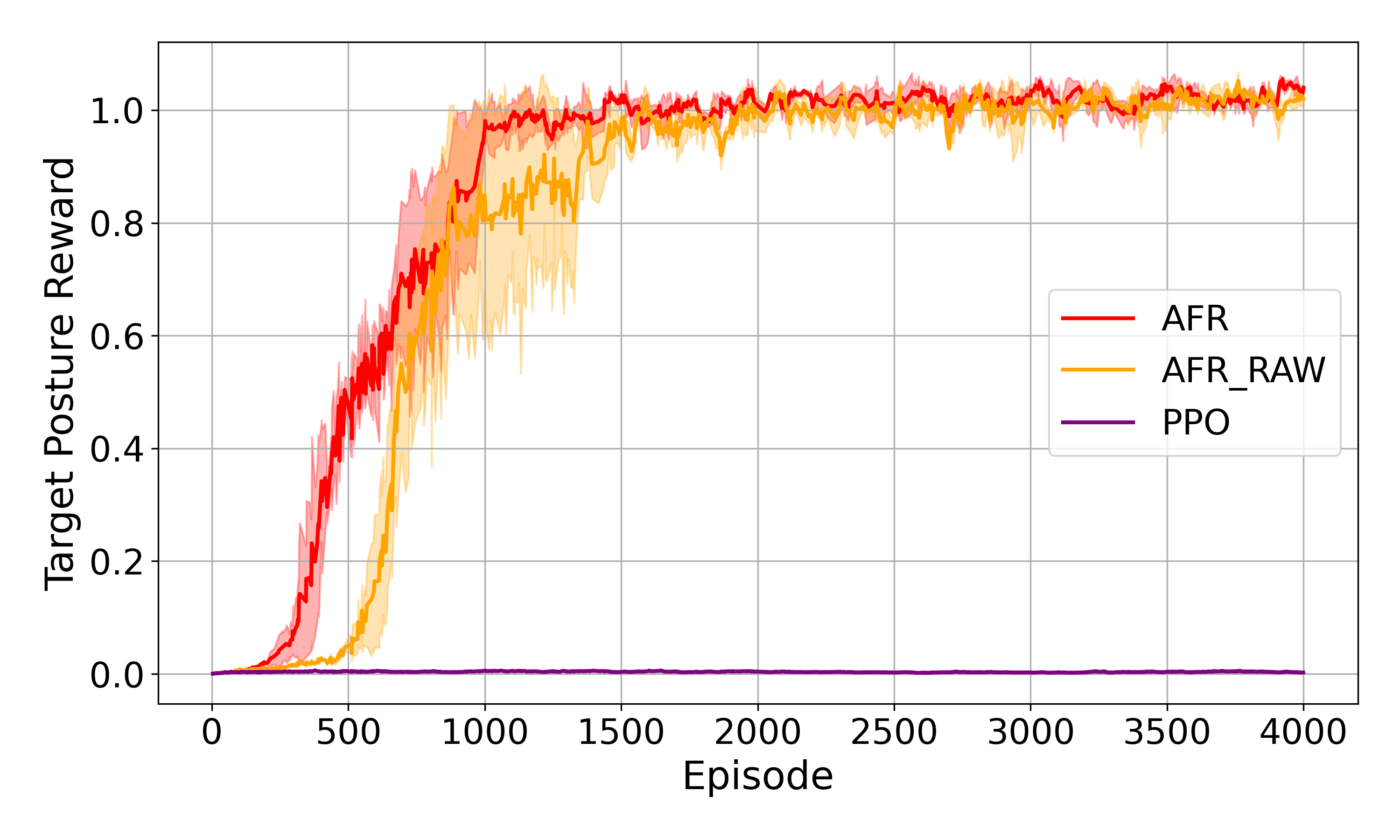}
   \caption{Target Posture Reward of AFR, AFR\textunderscore RAW and PPO.}
   \label{fig:target_posture_reward}
   \vspace{-0.5em}
\end{figure}

Fig. \ref{fig:overall_reward} and Fig. \ref{fig:target_posture_reward} illustrate the learning performance of AFR, AFR\textunderscore RAW, and PPO in terms of average rewards and target posture reward. AFR\textunderscore RAW inputs the uncompressed height map into the critic, while PPO uses single frame observations for both actor and critic. AFR significantly outperforms the others, likely due to its compressed elevation information in the critic and mass estimation in the actor. This mass prediction plays a crucial role in transferring learned behaviors across different platforms, such as from Go1 to other quadrupedal robots, enhancing environmental awareness and adaptability.

\subsubsection{Initial Posture and Joint Constraints}
We initially set a compact starting posture for safety during rolling. To balance safety constraints with effective exploration, we removed penalties for deviations from this posture, retaining only joint limit penalties, and introduced conditional rewards for specific joint positions when the robot's Euler angles indicate a normal standing posture. This approach maintains safety while allowing greater action exploration, facilitating the discovery of effective motion patterns for both rolling over and stable posture maintenance.



\subsubsection{Foot Contact Reward and Posture Balance}

We introduced a foot contact reward to encourage stable standing. However, high weights for this reward led to undesired ``frog squat" postures, highlighting the challenge of balancing stability and ideal posture in reward function design.

To address this, we fine-tuned the trade-off between foot contact rewards and posture maintenance rewards. This balance is crucial for guiding the robot to learn a standing posture that is both stable and upright, demonstrating the importance of careful reward function design in achieving desired behaviors.

\begin{figure*}[h]
   \centering
   \includegraphics[width=0.98\textwidth, height=0.98\textwidth]{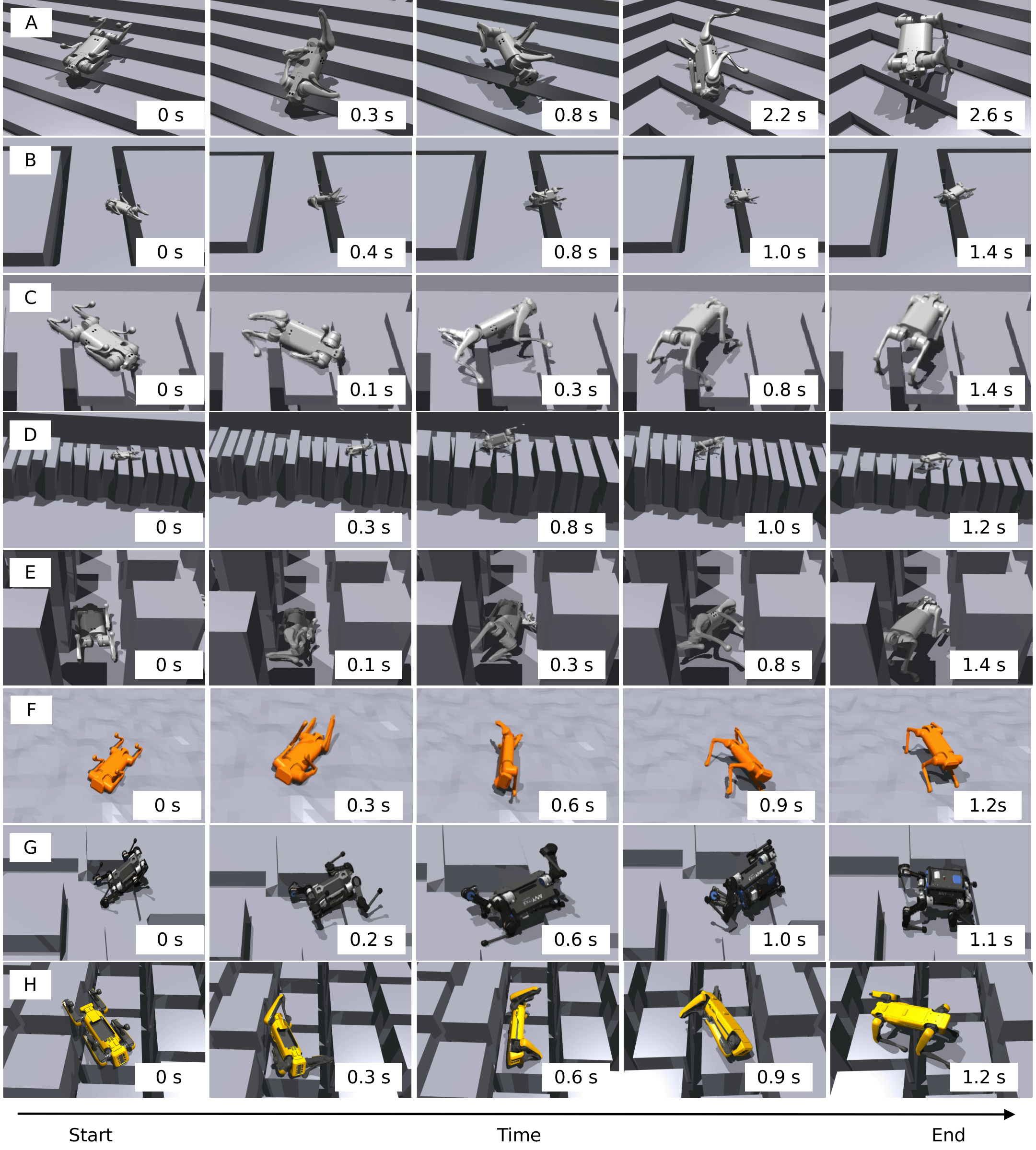}
   \caption{Simulation results of the Go1, Aliengo, Anymal B and Spot robots recovering from a prone position on uneven terrain. Each row (A-H) represents a different terrain: (A) Stairs, (B) Single Gaps, (C) Air Beams, (D) Beams, (E) Sparse Stones, (F) Slope, (G) Discrete Obstacles, and (H) Dense Stones. Columns show key phases in the recovery process across time steps.}
   \label{fig:sim_results}
\end{figure*}
\subsection{Robustness over challenging terrains}

\subsubsection{Torque Analysis during Stair Recovery}

\begin{figure}[h]
\centering
\includegraphics[width=0.45\textwidth,  trim=1 1 1 1,clip]{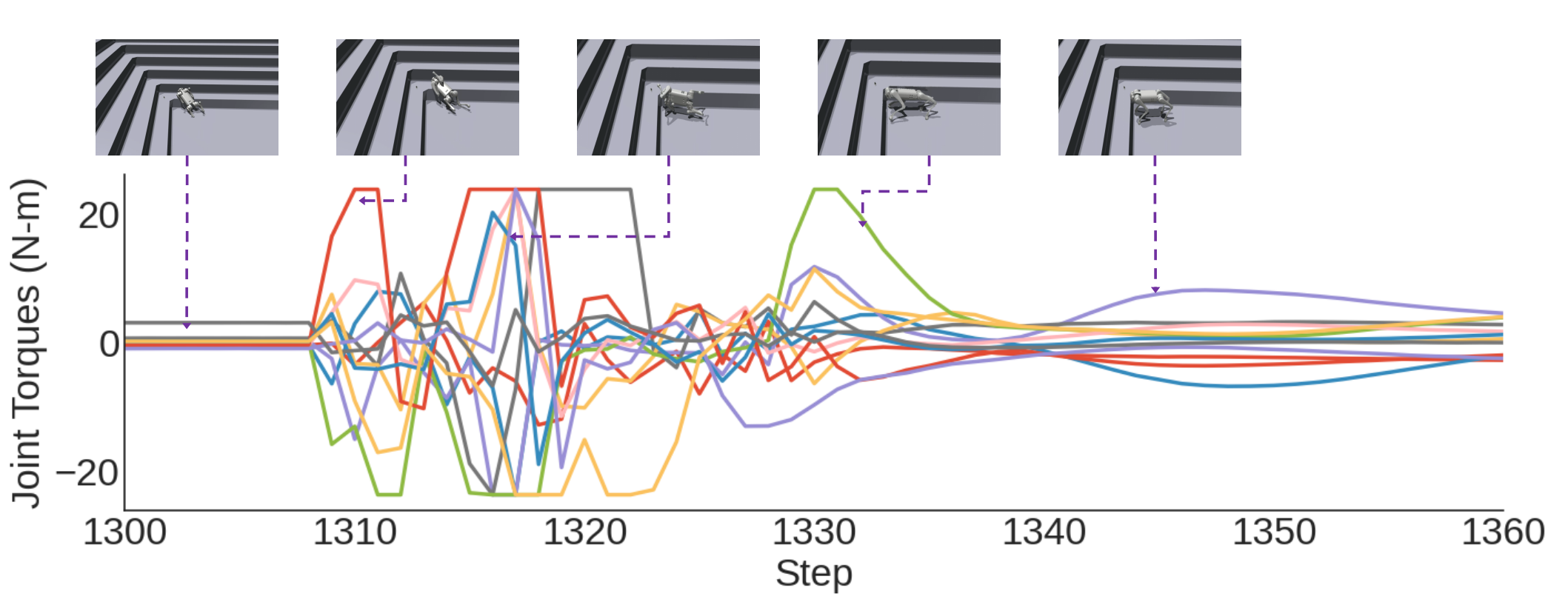}
\caption{Joint torque profiles during recovery on stairs.}
\vspace{-0.5em}
\label{fig:torque_analysis}
\end{figure}

To further understand the performance of our controller during challenging scenarios, we analyzed the joint torques during the recovery process on stairs. Fig. \ref{fig:torque_analysis} shows the torque profiles for each joint during a typical recovery sequence. 

As observed in Fig. \ref{fig:torque_analysis}, the AFR controller demonstrates efficient torque management across all joints. The hip joints experience the highest torque demands during the initial push-off phase, while the knee and ankle joints show more consistent torque patterns throughout the recovery process. This torque distribution indicates that our controller effectively utilizes the robot's mechanical structure to generate the necessary forces for recovery while minimizing unnecessary strain on the joints.

\subsubsection{Performance Comparison Across Terrains}
We evaluated our AFR approach against the PPO baseline across various terrains (Fig. \ref{fig:sim_results}), testing each method 50 times per terrain type. Table \ref{tab:terrain_comparison} presents key performance metrics: success rate (percentage of episodes achieving stable standing with varying payloads) and recovery time (average duration from supine to standing position). These metrics comprehensively assess the learned policies' effectiveness in diverse environments.

\vspace{-1.0em}
\begin{table}[htbp]
\caption{Performance of AFR and PPO across Different Terrains}
\centering
\vspace{-1.0 em}
\label{tab:terrain_comparison}
\begin{tabular}{lcccc}
\hline
\multirow{2}{*}{Terrain} & \multicolumn{2}{c}{Ours (AFR)} & \multicolumn{2}{c}{PPO} \\
\cline{2-5}
 & Success Rate & Time & Success Rate & Time \\
\hline
Slope & 98\% & 1.2s & 96\% & 1.2s \\
Discrete Obstacles & 86\% & 1.1s & 88\% & 1.7s \\
Stairs & 76\% & 2.6s & 62\% & 2.9s \\
Single Gaps & 62\% & 1.4s & 42\% & 2.5s \\
Air Beams & 60\% & 1.4s & 20\% & 2.6s \\
Beams & 64\% & 1.9s & 44\% & 2.3s \\
Sparse Stones & 92\% & 1.4s & 86\% & 1.5s \\
Dense Stones & 66\% & 1.2s & 52\% & 3.3s \\
\hline
\end{tabular}
\vspace{-1.5em}
\end{table}

\subsection{Sim-to-Sim Transfer}
To validate our policy's robustness and generalization capability, we performed sim-to-sim transfer from Isaac Gym to Gazebo, including tests on previously unseen terrain types.
\begin{figure}[h]
   \centering
   \includegraphics[width=0.40\textwidth, height=0.2\textwidth, trim=1 1 1 1,clip]{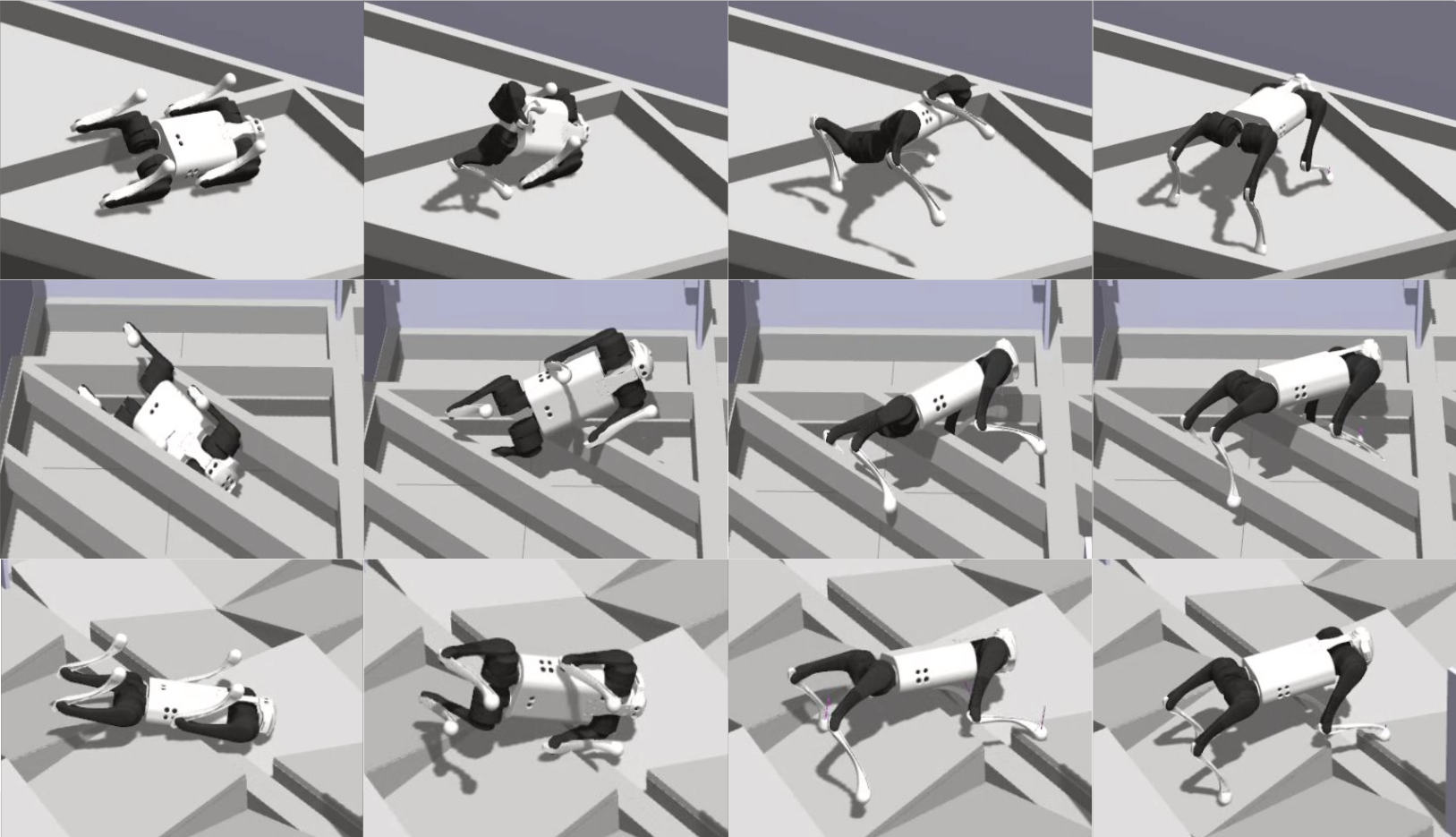}
   \caption{Fall Recoervy in Gazebo, demonstrating successful policy transfer.}
   \label{fig:gazebo_go1}
\end{figure}
Fig. \ref{fig:gazebo_go1} illustrates the recovery motion sequences in Gazebo environments. The policy trained in Isaac Gym was directly applied to Gazebo without fine-tuning. We conducted extensive trials on both familiar and novel terrain types in Gazebo, including randomly generated uneven surfaces (±25 cm height variations) and previously unseen structures like cantilever beams. Remarkably, the policy demonstrated no performance decline in the new environment.

The policy's consistent performance across simulators, especially on unseen terrains, validates our training approach and indicates readiness for real-world deployment. This successful transfer highlights the potential for smooth transition to physical robot implementations and underscores the value of multi-simulator validation in developing robust control policies for quadrupedal robots.





\section{CONCLUSION}

Our research introduces an innovative framework for developing resilient recovery strategies in quadrupedal robots, enabling them to regain stability across challenging terrains. Comprehensive simulations and experiments validate the efficacy of our approach, demonstrating successful adaptive pose restoration in quadrupedal systems.

Future work will integrate recovery policy with fall detection and locomotion, creating a unified system for seamless movement and swift recovery. We also plan to further deploy our controller on real robots with 
exteroceptive sensors (i.e. LIDAR) in challenging terrains to validate their robustness and applicability in extreme conditions.

\addtolength{\textheight}{-12cm}   







\printbibliography{}

\end{document}